\documentclass{article}
\usepackage{threeparttable}
\usepackage[export]{adjustbox}

\usepackage{microtype}
\usepackage{graphicx}
\usepackage{subcaption}
\usepackage{booktabs} 
\usepackage{amsmath}
\usepackage{booktabs,colortbl,xcolor,siunitx}
\usepackage{amssymb}
\usepackage{multirow,threeparttable}
\usepackage{makecell}
\usepackage{enumitem}
\definecolor{interiorred}{HTML}{B14434}
\definecolor{anchorgreen}{HTML}{426D3C}
\usepackage{threeparttable}
\newcommand{\isq}{\textcolor{interiorred}{\raisebox{0.15ex}{\rule{1.1ex}{1.1ex}}}}
\newcommand{\atri}{\textcolor{anchorgreen}{$\blacktriangle$}}

\definecolor{llamablue}{HTML}{406EAB}
\definecolor{mistralred}{HTML}{B14434}

\newcommand{\mkcirc}[1]{\textcolor{#1}{\raisebox{0.05ex}{$\bullet$}}}
\newcommand{\mksq}[1]{\textcolor{#1}{\raisebox{0.15ex}{\rule{1ex}{1ex}}}}
\newcommand{\mktri}[1]{\textcolor{#1}{$\blacktriangle$}}
\newcommand{\mkdia}[1]{\textcolor{#1}{$\blacklozenge$}}

\usepackage{hyperref}

\usepackage[accepted]{mlsys2025}

\makeatletter
\renewcommand{\mlsys@appearing}{\textit{Preprint.}}
\makeatother

\mlsystitlerunning{Anchor: Augmented Cross-Layer Recurrent State for Long-Context Retrieval in Test-Time Training}

\begin{document}

\twocolumn[
\mlsystitle{Anchor: Augmented Cross-Layer Recurrent State for Long-Context Retrieval in Test-Time Training}



\mlsyssetsymbol{equal}{*}

\begin{mlsysauthorlist}
\begin{mlsysauthorlist}
\mlsysauthor{Vimal William}{ua}
\mlsysauthor{Yuanzhe Li}{ua}
\mlsysauthor{Danella Zhao}{ua}
\mlsysauthor{Huanrui Yang}{ua}
\end{mlsysauthorlist}

\mlsysaffiliation{ua}{School of Electrical, Computer \& Software Engineering, University of Arizona, Tucson, AZ, USA}

\mlsyscorrespondingauthor{Huanrui Yang}{huanruiyang@arizona.edu }
\end{mlsysauthorlist}

\mlsyskeywords{Machine Learning, MLSys}

\vskip 0.3in

\begin{abstract}
Parameter-efficient linearization turns a pretrained autoregressive Transformer into an efficient model by hybridizing sliding-window attention (SWA) with a constant-size recurrent
memory. But does every layer's recurrent state contribute to long-context retrieval? Investigating an SWA and Test-Time Training (TTT) hybrid, we find that retrieval is sparse across depth,
concentrated in a few anchor layers, and that the remaining states hold similar information expressed through layer-specific transformations. We exploit both observations in
\textsc{Anchor}: TTT memory placed only at the anchor layers, and a cross-layer recurrent state written once and read by many layers, each through its own learned read map that realizes a
layer-specific stretch and rotation, along with demonstrated generalization to other hybrid attentions. On Llama-3.1-8B and Mistral-7B, \textsc{Anchor} delivers 2.3--3.1$\times$ higher recall and 6.7--8.9 points higher MMLU accuracy than LoLCATs with a
10$\times$ smaller recurrent state, retains 91--95\% of the teacher's MMLU accuracy, and cuts end-to-end latency by up to 4.3$\times$ relative to softmax attention.
\end{abstract}
]
\printAffiliationsAndNotice{~} 

\section{Introduction}

The self-attention mechanism at the core of large language models (LLMs) computes a similarity between every query and every preceding key, so both compute and memory grow with the sequence length. During autoregressive
decoding, the keys and values of all previous tokens are retained in a KV cache that grows linearly with context, requiring $\mathcal{O}(Nd)$ memory per layer for a sequence of length $N$ and head dimension $d$, while attention over the full sequence costs $\mathcal{O}(N^2d)$. This cache is what gives attention its defining property: every earlier token remains individually addressable, and recall over long contexts stays exact. It is also what makes long-context inference expensive. The cache must be held for the lifetime of a request and
re-read at every decoding step, so it, rather than arithmetic, bounds the batch size a server can sustain and therefore the throughput it can deliver. As context windows extend from thousands to hundreds of thousands of tokens, the
memory required to preserve exact recall becomes the dominant cost of deployment. This tension, between the growing state that gives attention its recall and the bounded state that makes inference affordable, \textbf{motivates architectures that retain long-range information in a memory whose size does not grow with the context}.

\begin{figure}[h]
  \centering
  \includegraphics[width=\columnwidth, right]{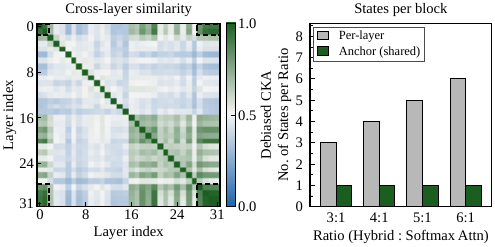}
  \caption{\textbf{Recurrent states across layers are redundant, so one shared state suffices.}
  \textbf{Left:} Debiased CKA between the TTT states of every pair of layers in Llama-3.1-8B,
  measured on WikiText for targets beyond the local window. The anchor layers
  ($\ell \in \{0, 1, 28, \dots, 31\}$, dashed boxes) are strongly aligned with one another
  (mean CKA $0.92$), whereas the interior layers are not ($0.49$).
  \textbf{Right:} Number of recurrent states per block as the ratio of hybrid to
  softmax-attention layers grows. A per-layer design keeps one state per hybrid layer;
  \textsc{Anchor} keeps a single shared state, reducing the count by $3$--$6\times$.}
  \label{fig:hero}
\end{figure}

A natural response is to impose an upper bound on the state. Recency-based and
importance-based architectures do so by restricting which past positions each
query may attend to: sliding window
attention~\cite{beltagy2020longformerlongdocumenttransformer} retains the $w$ most recent tokens, and sparse attention~\cite{deepseekai2025deepseekv32pushingfrontieropen} selects a fixed budget of earlier positions. The cache is then bounded at $\mathcal{O}(wd)$ and the per-token retrieval cost becomes independent of the sequence length,
restoring a predictable memory footprint at serving time. The bound, however, is enforced by discarding information rather than by compressing it. A token outside the retained set is no longer addressable at any later step, and no subsequent computation can recover it. The width $w$ therefore fixes the receptive field of the architecture, and recall falls away once the distance between a query and its target exceeds that width. Bounding the state exchanges one limitation for another: the memory cost no longer grows with context, but neither does the span over which the model can retrieve. 

The observed inefficiency in the recall accuracy due to the limitation in the receptive field motivates extending the receptive field through embedding a constant-sized recurrent state~\cite{chen2021scatterbrain, 3692070.3692524, arora2025simplelinearattentionlanguage, zhang2025lolcats, lan2025liger, william2026glide} to enable longer-context retrieval without a major increase in memory footprint and with the retrieval complexity of $\mathcal{O}(d^2 + wd)$. The recurrent state extends the span over which information is retained, but it remains an imperfect substitute for an explicit cache. Its capacity is limited by the low rank of the outer-product update, and because the state is written at
every step with no mechanism for forgetting, recent tokens progressively overwrite earlier ones. A line of work addresses these limitations by modifying the update rule rather than the state itself: gating and decay terms control how
much of the existing state is preserved at each step~\cite{yang2023gated}, and attentional-bias formulations replace the plain key-value association with an
objective optimized online~\cite{behrouz2025titans, oswald2026mesanet}, connecting the update to the in-context learning behavior observed in large language models~\cite{oswald2026mesanet}. These refinements improve how the state is written. The state itself, however, remains a single fixed-size summary maintained independently in every layer, and how much of it each layer actually requires has not been examined. This gap in understanding how long-term memory is distributed across depth leads us to ask: \textit{\textbf{does long-context retrieval require a recurrent state
in every layer and do states share information?}}

As a first step toward answering this question, we study the behavior of
recurrent states in a hybrid attention formulation on the associative recall
(AR) benchmark, and measure the similarity between the states of different
layers using centered kernel alignment (CKA) and cosine similarity. Two
observations follow (Figure~\ref{fig:hero}): long-context retrieval is sparse
across depth, concentrated in a small set of layers, and the states that carry
it hold closely related information. Together these motivate
\textbf{\textsc{Anchor}}, a recurrent state shared across layers rather than
maintained independently in each, which we instantiate as a family of
configurations spanning training and deployment. Our contributions are as follows:

\begin{enumerate}
    \item \textbf{Linearization for TTT states.} We extend the two-stage
    linearization recipe to Test-Time Training recurrent states, and show that
    supervising attention transfer with a next-token objective prevents the
    collapse that output matching alone produces (Section~\ref{sec:analysis}).

    \item \textbf{The anchor effect.} We find that long-context recall is
    concentrated in a small set of boundary layers: disabling the TTT update in
    the remaining $26$ layers leaves perplexity unchanged while shrinking the
    recurrent state by $5.3\times$ (Section~\ref{sec:analysis}).

    \item \textbf{Cross-layer recurrent state.} Layer states turn out to encode
    closely related information in different bases, which lets a single state
    serve many layers through learned per-layer read-maps $\chi_\ell(\cdot)$
    (Section~\ref{sec:cross_layer}).

    \item \textbf{\textsc{Anchor}.} The resulting family spans training and
    deployment and reaches $4.3\times$ higher end-to-end throughput compared to vanilla softmax attention at
    $16\times$ smaller recurrent state than per-layer hybrids, at comparable
    accuracy (Section~\ref{sec:results}).
\end{enumerate}

\section{Preliminaries}
We briefly review the attention mechanism, the core component of the autoregressive transformer architecture, and its impact on long-context inference; an extended discussion is provided in the Appendix~\ref{apd:prelm}.

\paragraph{Notations.} We denote by $N$ the sequence length, $L$ the number of layers, $H$ the number of heads, $d_{\text{model}}$ the model width, and $d = d_{\text{model}}/H$ the per-head dimension. Layers are indexed by
$\ell \in \{0,\dots,L-1\}$ and positions by $i$ or $t$. For an input $x_i \in \mathbb{R}^{d_{\text{model}}}$, the query, key, and value are $\mathbf{\hat{q}}_i = \mathbf{W_Q}^{\top}\mathbf{x}_i$, $\mathbf{\hat{k}}_i = \mathbf{W_K}^{\top}\mathbf{x_i}$, and $\mathbf{v_i} = \mathbf{W_V}^{\top}\mathbf{x_i}$; $o_i$ is the output of a single branch and $O_i$ that of the hybrid layer. $W$ is the
sliding-window size, $\mathcal{W}^{(c)}$ is the weights of recurrent state after chunk $c$, and $\boldsymbol{\chi}_\ell(\cdot)$ the learned read map at layer $\ell$. We write $\mathbf{\mathcal{A}}$ for cross-layer recurrent state memory, $g$ for an anchor group, $\boldsymbol{\theta}$
for the model parameters, $\mathrm{sg}[\cdot]$ for the stop-gradient operator, and $\mathcal{L}$ for objectives.

\paragraph{Self-Attention.} In\textit{ vanilla softmax attention}~\cite{vaswani2017attention}, for an input token representation $\mathbf{x}_i \in \mathbb{R}^{d_{\text{model}}}$, the query, key, and value are obtained as $\mathbf{\hat{q}}_i = \mathbf{W_{\text{Q}}}^{\top}\mathbf{\mathbf{x}}_i \in \mathbb{R}^{d_k}$, $\mathbf{\hat{k}}_i = \mathbf{W_{\text{K}}}^{\top}\mathbf{x}_i \in \mathbb{R}^{d_k}$, and $\mathbf{\hat{v}}_i = \mathbf{W_{\text{V}}}^{\top}\mathbf{x}_i \in \mathbb{R}^{d_v}$, with projection matrices $\mathbf{W_{\text{Q}}}, \mathbf{W_{\text{K}}} \in \mathbb{R}^{d_{\text{model}} \times d_k}$ and $\mathbf{W_{\text{V}}} \in \mathbb{R}^{d_{\text{model}} \times d_v}$, replicated across $H$ heads with $d_k = d_v = d_{\text{model}}/H$. Attention predicts the distribution over the next token auto-regressively by computing the similarity between the current query $\mathbf{\hat{q}}_i$ and the keys of all preceding positions, retained together with their values as the KV cache $\{(\mathbf{k}_j, \mathbf{v}_j)\}_{j \leq i}$. The weight assigned to position $j$ is $\boldsymbol{\alpha}_{ij} = \operatorname{sim}(\mathbf{\hat{q}}_i, \mathbf{\hat{k}}_j) / \sum_{j'=0}^{i} \operatorname{sim}(\mathbf{\hat{q}}_i, \mathbf{\hat{k}}_{j'})$ with $\operatorname{sim}(\mathbf{\hat{q}}, \mathbf{\hat{k}}) = \exp(\mathbf{\hat{q}}^{\top}\mathbf{\hat{k}}/\sqrt{d_k})$, giving $\boldsymbol{\alpha}_i \in \mathbb{R}^{i+1}$ with $\sum_j \boldsymbol{\alpha}_{ij} = 1$ and the layer output $\mathbf{o}_i = \sum_{j \leq i} \boldsymbol{\alpha}_{ij}\mathbf{\hat{v}}_j \in \mathbb{R}^{d_v}$. \textit{Because the exponential similarity admits no finite-dimensional factorization, the sum over $j$ cannot be accumulated into a fixed-size state: the full cache must be retained, costing $\mathcal{O}(\text{N}^2 d)$ compute and $\mathcal{O}(\text{N}d)$ for a sequence of length $\text{N}$. }

\paragraph{In-Context Recall.} Because the exponential similarity cannot be factorized, the cache must be kept in full, and softmax attention retains exact access to every earlier token. This is what gives it strong recall over long contexts, but it comes at two costs that both grow with the sequence length $N$: the weighted sum $\sum_{j \leq i} \alpha_{ij}\mathbf{v}_j$ is taken over all preceding positions at every step, and the cache $\{(\mathbf{k}_j, \mathbf{v}_j)\}_{j \leq i}$ itself expands without bound. Alternative formulations address this by capping how much of the past each query may access. Sliding-window attention~\cite{beltagy2020longformerlongdocumenttransformer} restricts the sum to the $W$ most recent positions, and sparse attention~\cite{child2019generatinglongsequencessparse, deepseekai2025deepseekv32pushingfrontieropen} selects a subset of earlier positions under a fixed budget; \textit{in both cases the cache is bounded at $\mathcal{O}(W d)$ rather than $\mathcal{O}(N d)$. The overhead is removed, but so is access to anything outside the retained window, and recall on long-context retrieval degrades accordingly.}

\paragraph{Recurrent State as Long-term Memory.} The observed degradation in recall accuracy on long-context retrieval can be
compensated either by augmenting self-attention with a constant-size recurrent state~\cite{arora2025simplelinearattentionlanguage, zhang2025lolcats, lan2025liger, william2026glide} that serves as long-term memory, or by replacing self-attention with such a state entirely. One way to obtain this state is to linearize the attention sum $\sum_{j=0}^{i}\mathrm{sim}(\mathbf{\hat{q}}_i, \mathbf{\hat{k}}_j)$ with a separable kernel feature map $\phi$, which yields $\phi(\mathbf{\hat{q}}_i)\sum_{j=0}^{i}\phi(\mathbf{\hat{k}}_j)^{\top}$. The attention output $\mathbf{O}_i$ under linearized attention~\cite{katharopoulos2020transformersrnnsfastautoregressive} is then $\mathbf{O}_i = \frac{\phi(\mathbf{\hat{q}}_i)\mathbf{S}_i}{\phi(\mathbf{q}_i)^{\top}\mathbf{Z}_i}$, where $\mathbf{S_i} = \sum_{j=0}^{i}\phi(\mathbf{k_j})^{\top}\mathbf{v_j} \in \mathbb{R}^{d \times d}$ and the normalization factor is $\mathbf{Z_i} = \sum_{j=0}^{i}\phi(\mathbf{k_j}) \in \mathbb{R}^{d}$. The state accumulates a summary of the context through KV binding and is updated recurrently as $\mathbf{S}_i = \mathbf{S}_{i-1} + \phi(\mathbf{k}_i)^{\top}\mathbf{v}_i$. In this form, the state is written at every step with no mechanism for forgetting, which limits its performance, and the
effect can be reduced by introducing a gating mechanism into the update rule~\cite{yang2023gated, yang2025gated,
hatamizadeh2026gateddeltanet2decouplingerase, oswald2026mesanet}. \textit{Beyond the update rule, linearized attention remains less expressive than softmax attention because the state is low-rank, and this limited expressiveness is reflected in poor recall accuracy.
}

\paragraph{Neural Network as Long-term Memory.} Motivated by the need to improve the expressiveness of the recurrent state and
to support online adaptation to the context, Test-Time
Training~\cite{sun2024learning, sun2025learning} approximates the recurrent state with a neural network trained on a key-value binding objective $\ell\big(\mathbf{\mathcal{M}}_{t-1}; x_t\big)$~\cite{behrouz2025titans, zhang2026testtime} at step $t$. The state is updated by a gradient descent step
on this key-value association, which is referred to as the \textit{attentional bias}, and is read out as $\mathbf{\mathcal{M}}_t(\mathbf{\hat{q}}_t)$. Subsequent work refines the objective used to encode the context: Atlas~\cite{behrouz2026atlas} encodes a
sampled context window into the recurrent state rather than relying on the key-value association alone, and a recent
formulation~\cite{DBLP:journals/corr/abs-2512-23675} casts the state update as a next-token prediction objective. \textit{Despite its expressiveness, the inner-loop gradient modifications introduce a state-update overhead, and the reliance on expensive pre-training limits adaptivity.}

\paragraph{Linearized Language Modeling.} Given an input sequence $x_{\le t}=(x_1,\dots,x_t)$,
language modeling trains the model to predict the next token $x_{t+1}$ by minimizing the
cross-entropy between the predicted distribution and the true token, $\mathcal{L}_{\mathrm{CE}} = -\frac{1}{B(T-1)}\sum_{b}\sum_{t=1}^{T-1}\log p_\theta(x_{b,t+1}\mid x_{b,\le t})$, where $B$ and $T$ denote the batch size and sequence length. Linearization converts a pretrained
Transformer into an efficient model by replacing its softmax attention with a sub-quadratic
student layer while reusing the remaining weights~\cite{zhang2025lolcats,lan2025liger}. It first
performs attention transferring through distillation, matching the output of each student
layer to the softmax-attention output of the frozen teacher through
$\mathcal{L}_{\mathrm{MSE}} = \frac{1}{N}\sum_{\ell=1}^{L}\frac{1}{BTd}\sum_{b,t}
\lVert \mathbf{o}_{b,t}^{(\ell)} - \mathrm{sg}[\mathbf{a}_{b,t}^{(\ell)}]\rVert_2^2$, where $d$ is the
head dimension, $L$ the number of layers and $\mathrm{sg}[\cdot]$ the stop-gradient operator, and
then fine-tunes the model on $\mathcal{L}_{\mathrm{CE}}$ with parameter-efficient
adapters. \textit{Linearization preserves pretrained knowledge efficiently, yet its recall remains
limited by the expressiveness of the sub-quadratic layer.}

\begin{figure*}[thb!]
  \centering
  \includegraphics[width=0.90\textwidth]{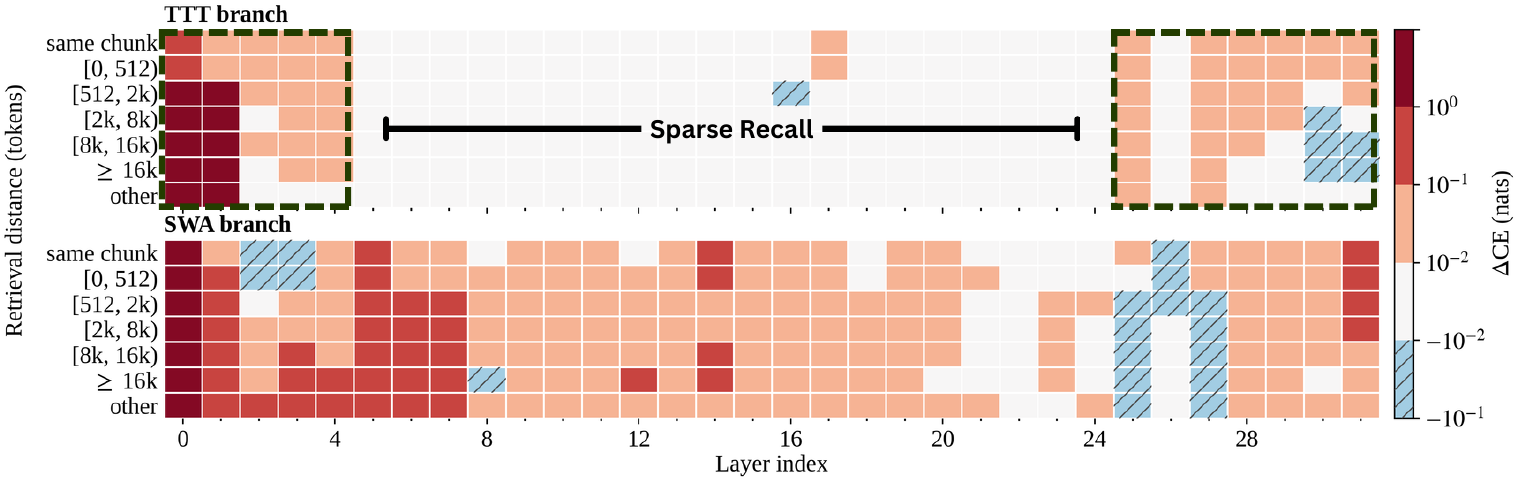}
  \caption{\textbf{Layer-wise contribution of each branch to long-context
retrieval.} $\Delta\mathrm{CE}$ (nats) on AR-32k when the TTT (top) or SWA
(bottom) branch is removed at a single layer, broken down by the distance
between a query and its target. Warm cells indicate degradation, that is, a
larger increase in loss when the branch is removed. The TTT branch contributes
almost nothing across the interior layers ($\ell \in \{2,\dots,27\}$) but
dominates at the boundary layers ($\ell \in \{0,1,28,\dots,31\}$), whereas SWA
contributes broadly at every depth. The color scale is symmetric-log with a
linear region below $0.01$, and both panels share one scale.}
  \label{fig:dce-heatmap}
\end{figure*}

\section{Methodology}
In this section, we introduce our method \textbf{\textsc{Anchor}}, an effective linearization of LLMs through the combination of Sliding Window Attention (SWA) for strong short context and a Test-Time Training (TTT) module for efficient longer context memory. We initially present our improved linearization framework and an empirical study on the window attention and TTT module to understand the internal information processing (Sec.\ref{sec:analysis}). We position our long-term memory state unification method, driven by our findings, that longer-context retrieval is sparser and shares related information across layers.

\begin{table}[h!]
\caption{\textbf{Ablation of the test-time-training (TTT) module on
long-context performance.} Disabling the TTT update in the anchor layers
($\ell \in \{0,1,28,29,30,31\}$) raises out-of-domain perplexity
on WikiText by $48\times$, while disabling it in the interior layers
($\ell \in \{2,\dots,27\}$) leaves perplexity unchanged and shrinks the
persistent state by $5.3\times$. SWA remains active in all layers.}
\label{tab:ttt-layers}
\vskip 0.15in
\begin{center}
\begin{small}
\setlength{\tabcolsep}{11pt}
\renewcommand{\arraystretch}{1.15}
\begin{tabular}{l S[table-format=3.2] S[table-format=3.1]}
\toprule
\textbf{Configuration} & {\textbf{PPL} $\downarrow$} & {\textbf{State (MB)} $\downarrow$} \\
\midrule
Llama-3.1-8B (baseline)  & 7.36   & {--}  \\
\midrule
\addlinespace[2pt]
\rowcolor{green!8}
Non-unified state        & 10.19  & 201.3 \\
\midrule
\quad $-$ \textbf{Anchor} TTT     & 487.56 & 163.6 \\
\quad $-$ Interior TTT   & 10.01  & 37.7  \\
\bottomrule
\end{tabular}
\end{small}
\end{center}
\vskip -0.1in
\end{table}

\subsection{Empirical Analysis} \label{sec:analysis}

\begin{figure}[h!]
  \centering
  \includegraphics[width=\columnwidth]{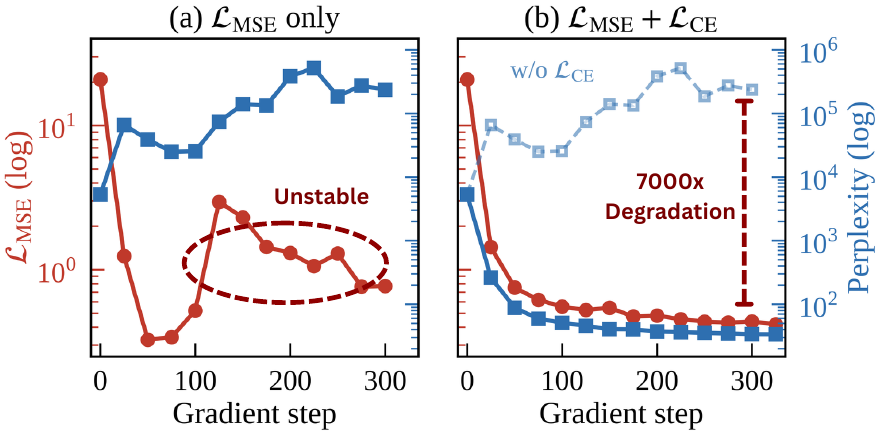}
  \caption{\textbf{Improved Attention Transfer in Stage-1 Linearization.} Held-out attention-transfer loss $\mathcal{L}_{\mathrm{MSE}}$ (left) and perplexity (right) during Stage~1. (a) Training on $\mathcal{L}_{\mathrm{MSE}}$ alone matches the teacher's attention outputs closely, reducing the loss by over $25\times$, yet perplexity rises from 5.3K to 236K: the student reproduces each layer's output while losing the ability to model language. (b) Adding next-token supervision $\mathcal{L}_{\mathrm{CE}}$ resolves this, lowering both losses together and reaching a perplexity of 34, which provides a stable initialization for the TTT parameters.}
  \label{fig:stage1}
\end{figure}

\paragraph{Improved Linearization Framework.} Existing SWA--TTT hybrids are trained from scratch~\cite{zhang2026testtime, behrouz2026atlas, behrouz2025titans}, an approach whose cost scales poorly with model size. We therefore adopt the
two-stage linearization of LoLCATs~\cite{zhang2025lolcats} and use its attention-transfer stage to initialize the TTT module. This alone is insufficient: stage 1 drives the mean squared error (MSE) down rapidly while model quality collapses (Figure~\ref{fig:stage1}), indicating that the MSE
objective alone is a poor proxy for the behavior we wish to transfer. We address this by casting attention transfer as next-token prediction~\cite{DBLP:journals/corr/abs-2512-23675} regularized by the MSE term, which prevents the catastrophic degradation. We define the resulting
attention-transfer objective $\mathcal{L}_{\text{AT}}$ as

\begin{equation}
\label{eq:at}
\mathcal{L}_{\text{AT}} \;=\; \lambda\,\mathcal{L}_{\text{MSE}}
\;+\; w_{\text{LM}}\,\mathcal{L}_{\text{CE}},
\end{equation}
%

where $\mathcal{L}_{\text{MSE}} = \frac{1}{N}\sum_{\ell=1}^{L}\frac{1}{BTd} \sum_{b,t}\bigl\|\mathbf{o}^{(\ell)}_{b,t}-\mathrm{sg}\bigl[
\mathbf{a}^{(\ell)}_{b,t}\bigr]\bigr\|_2^2$ pulls the output $\mathbf{o}^{(\ell)}_{b,t}$ of each hybrid layer toward the
softmax-attention output $\mathbf{a}^{(\ell)}_{b,t}$ of the frozen teacher at the same layer and position, and
$\mathcal{L}_{\text{CE}} = -\frac{1}{B(T-1)}\sum_{b,t}\log
p_\theta\bigl(x_{b,t+1}\mid x_{b,\le t}\bigr)$
is the next-token cross-entropy over $t = 1,\dots,T-1$. Here $B$, $T$, $d$, and $N$ denote the batch size, sequence length, head dimension, and number of layers, respectively; $\mathrm{sg}[\cdot]$ is the stop-gradient operator, which
confines gradients to the student branch; and $\lambda$ and $w_{\text{LM}}$ control the trade-off between per-layer output matching and next-token prediction. Supervising attention transfer with $\mathcal{L}_{\text{CE}}$
stabilizes Stage~1 and yields an effective initialization of the TTT parameters. Stage~2 then applies LoRA finetuning on a long-context corpus (PG-19) whose sequences exceed the TTT chunk size and therefore engage the inner loop, providing a cost-efficient adaptation of the TTT module that
supports in-context memorization. Together, the two stages recover a linearized
model whose performance is close to that of the softmax-attention baseline
(Table~\ref{tab:ttt-layers}).

\paragraph{Anchor-Effect in Test-Time Training Branch.} To understand what the TTT module actually does in the hybrid architecture, we run a layer-wise ablation on the Associative Recall (AR) benchmark~\cite{arora2024zoology}. At a single layer $\ell$ we remove one branch, either TTT or SWA, and measure how much the cross-entropy rises relative to the unablated model. We report this separately for each range of distances between a query and its target:

\begin{equation}
\Delta\mathrm{CE}_k^{(\ell)} = \frac{1}{|\mathbf{\mathcal{V}_k}|}\sum_{i\in\mathcal{V}_k}
  \Big[\bar{\ell}^{(i)}_k(\boldsymbol{\theta\setminus\ell}) - \bar{\ell}^{(i)}_k(\boldsymbol{\theta})\Big],
\end{equation}

where $\mathcal{V}_k$ is the set of evaluation examples whose retrieval distance
falls in bucket $k$, and $\theta\setminus\ell$ denotes the parameters with the branch at layer $\ell$ removed. A large $\Delta\mathrm{CE}_k^{(\ell)}$ means that the removed branch was carrying retrieval information at that depth and
distance which no other component supplies. Measured this way, the two branches behave very differently (Figure~\ref{fig:dce-heatmap}). SWA contributes at every depth and across all retrieval distances. TTT does not: its contribution is
concentrated in a small set of boundary layers and is negligible everywhere in between ($\ell \in \{2,\dots,27\}$). We call this concentration the \emph{anchor effect} and the layers that show it \emph{anchor layers}, and Table~\ref{tab:ttt-layers} confirms that these few layers carry the load. Disabling the TTT update in the anchor layers raises out-of-domain perplexity on
WikiText from $10.19$ to $487.56$, a factor of $48$, whereas disabling it in all $26$ interior layers leaves perplexity where it was ($10.01$) and shrinks the persistent state from $201.3$\,MB to $37.7$\,MB, a factor of $5.3$. The interior
TTT state is therefore not just redundant: it can be dropped at no cost in quality, and it accounts for most of the model's state budget.

\begin{figure}[h]
  \centering
  \includegraphics[width=\columnwidth]{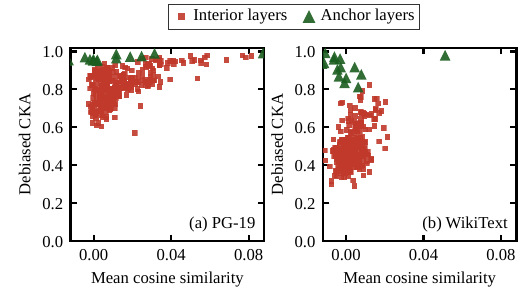}
  \caption{\textbf{Representational similarity of TTT states} on (a) PG-19 and
(b) WikiText. Each point is one layer pair, plotted as debiased CKA against
mean cosine similarity on cross-window targets, that is, those whose source
lies beyond the local window. Interior pairs (\isq, $\ell \in \{2,\dots,27\}$)
are drawn separately from anchor pairs (\atri,
$\ell \in \{0,1,28,\dots,31\}$). Anchor pairs are more strongly aligned than
interior pairs on both corpora ($0.97$ versus $0.82$ on PG-19, $0.92$ versus
$0.49$ on WikiText). High CKA with near-zero cosine indicates closely related
information encoded in different bases.}
  \label{fig:sim_check}
\end{figure}


\begin{figure*}[t]
\centering
\includegraphics[width=0.90\textwidth]{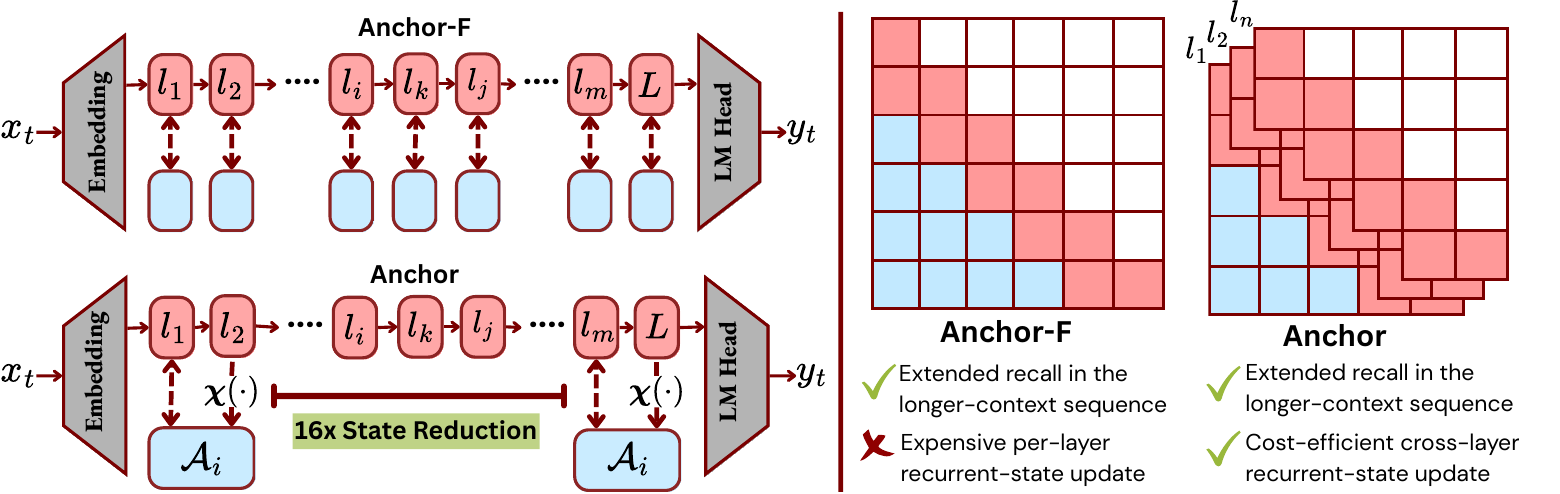}
\caption{\textbf{Anchor architecture and recurrent-state coverage.}
\emph{Left:} Anchor-F maintains a private recurrent state in every layer,
while Anchor keeps one shared state per anchor group and removes the
interior states entirely, reducing recurrent state by $16\times$
($201.0 \rightarrow 12.6$\,MB at 32k context). Each reader accesses the
shared state through its own learned read map $\chi(\cdot)$.
\emph{Right:} coverage over query (rows) and key (columns) positions.
Green marks the local sliding window, red the span served by the
recurrent state. Anchor-F requires one such state per layer; Anchor
serves layers $l_1,\dots,L$ from a single shared state.}
\label{fig:anchor}
\end{figure*}

\paragraph{Shared Long-term Memory Across Layers.} Since retrieval concentrates in the anchor
layers, we next ask whether these layers store distinct information or largely the same. We
compare the TTT states of every pair of layers on targets whose source lies beyond the local
window, using two complementary measures: debiased centered kernel alignment
(CKA)~\cite{kornblith2019similarity}, which is invariant to rotations and isotropic scaling and thus
detects shared information regardless of the coordinate frame, and mean cosine similarity, which
compares representations coordinate by coordinate and is high only when two layers share the same
frame. The two measures disagree sharply (Figure~\ref{fig:sim_check}). CKA between anchor layers
(\atri) approaches unity on both corpora, exceeds that of interior layers (\isq), lies far above a
shuffled-label baseline, and remains high out of domain, yet the cosine similarity of the same
pairs stays close to zero. \textit{Layers therefore encode closely related information in
layer-specific frames.} This has two consequences: keeping a separate state in every layer is
largely redundant, yet the states cannot simply be reused, since a reader would interpret the
shared content in the wrong frame. A single shared state can thus serve several layers only if each reader applies its own transformation, motivating the learned read maps introduced below
(Appendix~\ref{apd:readmap_long}).

\subsection{Anchor: Cross-Layer Recurrent State} \label{sec:cross_layer}
We present our method, \textbf{\textsc{Anchor}}, a cross-layer Test-Time Training
(TTT)-based memory module that extends the receptive field of Sliding Window
Attention (SWA) for longer-context inference through targeted adaptation. The
method is built on two observations: \textbf{1)} long-context retrieval is
reinforced mainly by the early and later TTT modules in the hybrid attention
(SWA \& TTT), and \textbf{2)} there is a representational relation between the
TTT states across layers, with a higher concentration in the early and later
modules. Based on these motivating observations, we design a cross-layer TTT
module that can be shared by multiple Sliding Window Attentions (SWA). Each
cross-layer module has a single writer and multiple readers, where every reader
applies its own basis matrix as a read-map, and it can be defined as,

\begin{equation}
o_\ell \;=\; \boldsymbol{\chi}_\ell\!\big(f_{W^{(c)}}(\mathbf{\hat{q}_\ell})\big),
\end{equation}

where $\mathbf{\hat{q}_\ell}$ is the query at layer $\ell$, $f_{W^{(c)}}$ is the shared
memory with state $\mathbf{W^{(c)}}$ which reads a context chunk $c$, and $\boldsymbol{\chi_\ell}(\cdot)$ is a learnable read-map
for the multiple readers across layers, which operate on a single state update
mechanism. The layer that performs the state update carries a reader initialized
with the identity matrix ($\chi = \mathbf{I}$), since the reader in that
layer shares the same basis as the writer and the state update rule can be defined as, $\mathbf{W^{(c)}} = \mathbf{W^{(c-1)}} - \eta\,\!\Big(
\nabla_{\!W}\,\mathcal{L}\mathbf{\big(W^{(c-1)}};\,
\mathbf{\hat{k}_{c-1}},\, \mathbf{\hat{v}_{c-1}}\big)\Big),$ Where the state update mechanism models the attentional bias through the objective function ($\mathcal{L}\mathbf{\big(W^{(c)}};\,
\mathbf{\hat{k}_{c}},\, \mathbf{\hat{v}_{c}}\big)$) in which the association within the KV-binding expresses as $\mathcal{L}(\mathbf{W}; \mathbf{\hat{k}}, \mathbf{\hat{v}})
\;=\; \tfrac{1}{2}\big\| f_W(\mathbf{\hat{k}}) - \mathbf{\hat{v}} \big\|_2^2$, which is an euclidean norm of estimating value ($f_W(\mathbf{\hat{k}})$) and actual value ($\mathbf{\hat{v}}$) at the instance of given chunk $c$.  Evaluating the hybrid across context
lengths, we find that \textbf{\textsc{Anchor}} substantially improves in-context retrieval once
the relevant content falls outside the window, whereas in the short-context regime, where SWA
alone suffices, the cross-layer state adds a small amount of noise. Its contribution should
therefore depend on whether the input requires memory beyond the window. We introduce a
learned, input-dependent gate on the readout of the cross-layer recurrent state:
\begin{equation}
  \mathbf{O}_t = \mathbf{O}_t^{\mathrm{SWA}} + \boldsymbol{\gamma}_t \odot \mathbf{O}_t^{\mathrm{TTT}},
\end{equation}
where $\gamma_{t,h} = \sigma(\mathbf{w}_h^\top\mathbf{x}_t + b_h)\in(0,1)$ for head $h$, with
learnable $\mathbf{w}_h \in \mathbb{R}^{d_{\mathrm{model}}}$ and $b_h \in \mathbb{R}$, is broadcast
across the head's value dimension. As $\gamma_{t,h}\to0$ the layer reduces to sliding-window
attention, and larger values admit more of the long-range readout while SWA retains the local
context. \textit{The gate costs one vector--matrix product per token and leaves the cost of the hybrid
layer essentially unchanged.}

\paragraph{Efficient Deployment Configuration.} The interior TTT branch is dispensable at inference, but not during fine-tuning. Training without it from the outset causes the loss to oscillate rather than converge smoothly, whereas the same configuration with the interior branch retained trains stably under identical settings. A plausible explanation lies in
how the gradient reaches the shared anchor states. With the interior branch present, each interior layer contributes its own readout gradient. Once it is removed, the entire long-range signal must instead flow through the anchor groups, so a
shared state $W^{(c)}_{g}$ accumulates $\sum_{\ell \in g} \partial\mathcal{L}/\partial o_\ell$ from every reader in its
group $g$, its effective step size grows with $|g|$, and the number of independent states falls from $L{=}32$ to $|\mathcal{A}|{=}2$. We therefore adopt a deployment-aware configuration: the interior branch is retained throughout Stage-1 and Stage-2 fine-tuning, where it stabilizes optimization and yields a smooth attention transfer, and is discarded once the model is deployed. The recurrent state shrinks by roughly $16\times$ between the training and
deployment configurations. \textit{The interior memory is therefore a plug-in component: present when
training needs it, absent when it only costs state.}

\section{Experiments}

In this section we describe the experimental setup used to validate
\textbf{\textsc{Anchor}} on language modeling and recall-intensive tasks. We
assess quality on downstream common-sense and recall benchmarks, and efficiency
along three axes: the reduction in recurrent state, end-to-end latency across a
range of sequence lengths, and the FLOPs required for the state update in the
TTT branch. Together, these show that \textsc{Anchor} preserves recall quality
while reducing both the memory and the compute cost of the recurrent state.

\begin{table*}[t]
\centering
\begin{threeparttable}
\caption{\textbf{Zero-shot downstream reasoning and recall performance.} Accuracy of the fine-tuned \textsc{Anchor} variants on six common-sense reasoning benchmarks (LAMBADA, PIQA, HellaSwag, WinoGrande, ARC-e, ARC-c) and three recall-intensive benchmarks (SWDE, FDA, SQuAD), compared against the softmax-attention baseline. \textsc{Anchor}-F gives every sliding window its own TTT module; \textsc{Anchor}-I shares one TTT module across the initial and final layers while the interior layers keep separate modules; \textsc{Anchor} shares one module across the anchor layers and removes the interior modules entirely. \textbf{(k/s) - K Token/sec}; measured at N = $8\text{K}$ and B = $19$.}
\label{tab:lm-eval}
\small
\setlength{\tabcolsep}{4pt}
\renewcommand{\arraystretch}{1.15}
\begin{tabular}{@{}l l >{\columncolor{green!15}}r >{\columncolor{green!15}}r cccccc >{\columncolor{green!15}}c ccc >{\columncolor{green!15}}c@{}}
\toprule
& & \multicolumn{1}{c}{\textbf{State}} & \multicolumn{1}{c}{\textbf{Tput}}
& \multicolumn{7}{c}{\textbf{Commonsense}}
& \multicolumn{4}{c}{\textbf{Recall}} \\
\cmidrule(lr){5-11}\cmidrule(l){12-15}
\textbf{Model} & \textbf{Config} & (MB)\,$\downarrow$ & (k/s)\,$\uparrow$
& LMB. & PIQA & Hella. & Wino. & ARC-e & ARC-c & Avg.
& SWDE & FDA & SQuAD & Avg. \\
\midrule
\multirow{6}{*}{LLaMA}
& Baseline          & --- & $1.73$ & $75.28$ & $81.12$ & $79.04$ & $73.64$ & $81.52$ & $53.50$ & \underline{$\mathbf{74.02}$} & $91.81$ & $81.03$ & $53.82$ & \underline{$\mathbf{75.55}$} \\
& SWA               & --- & \underline{$\mathbf{8.32}$} & $75.00$ & $80.47$ & $78.45$ & $73.09$ & $81.10$ & $52.73$ & $73.47$ & $27.72$ & $12.43$ & $48.66$ & $29.60$ \\
& LoLCATs           & $129.0$ & $3.18$ & $69.09$ & $81.61$ & $78.56$ & $73.95$ & $82.87$ & $55.20$ & $73.55$ & $18.99$ & $1.72$ & $32.57$ & $17.76$ \\
\cmidrule(l){2-15}
& \textbf{\textsc{Anchor}}-F & $201.0$ & $3.41$ & $74.21$ & $80.36$ & $76.63$ & $72.69$ & $80.56$ & $49.40$ & \underline{$72.31$} & $50.50$ & $18.60$ & $60.46$ & \underline{$43.19$} \\
& \textbf{\textsc{Anchor}}-I & $176.0$ & $3.66$ & $73.04$ & $79.16$ & $77.37$ & $72.69$ & $79.59$ & $47.44$ & $71.55$ & $48.51$ & $19.24$ & $54.02$ & $40.59$ \\
& \textbf{\textsc{Anchor}}   & \underline{$\mathbf{12.6}$} & \underline{$7.25$} & $72.27$ & $78.35$ & $76.49$ & $73.88$ & $79.17$ & $47.27$ & $71.24$ & $49.86$ & $24.32$ & $50.50$ & $41.56$ \\
\midrule
\multirow{6}{*}{Mistral}
& Baseline          & --- & $1.65$ & $75.61$ & $80.58$ & $81.10$ & $73.56$ & $80.89$ & $50.34$ & \underline{$\mathbf{73.68}$} & $89.83$ & $71.42$ & $51.41$ & \underline{$\mathbf{70.88}$} \\
& SWA               & --- & \underline{$\mathbf{8.44}$} & $75.98$ & $80.79$ & $81.33$ & $75.37$ & $80.01$ & $47.87$ & $73.56$ & $33.57$ & $12.34$ & $53.85$ & $33.25$ \\
& LoLCATs           & $129.0$ & $3.05$ & $69.80$ & $81.61$ & $80.35$ & $74.03$ & $82.11$ & $53.07$ & $73.50$ & $11.07$ & $0.09$ & $23.32$ & $11.49$ \\
\cmidrule(l){2-15}
& \textbf{\textsc{Anchor}}-F & $201.0$ & $3.26$ & $74.73$ & $80.85$ & $80.73$ & $72.93$ & $79.59$ & $48.81$ & \underline{$72.94$} & $48.15$ & $15.06$ & $52.92$ & \underline{$38.71$} \\
& \textbf{\textsc{Anchor}}-I & $176.0$ & $3.52$ & $73.88$ & $80.30$ & $80.06$ & $73.64$ & $79.17$ & $47.78$ & $72.47$ & $45.54$ & $13.88$ & $51.07$ & $36.83$ \\
& \textbf{\textsc{Anchor}}   & \underline{$\mathbf{12.6}$} & \underline{$7.15$} & $74.60$ & $80.69$ & $80.69$ & $73.09$ & $78.70$ & $47.78$ & $72.59$ & $44.82$ & $12.25$ & $49.53$ & $35.53$ \\
\bottomrule
\end{tabular}
\end{threeparttable}
\end{table*}

\subsection{Experimental Setup}
\paragraph{Implementation Details.} We build our implementation on Liger~\cite{lan2025liger} and
LoLCATs~\cite{zhang2025lolcats}, and initialize the TTT module from the
reference implementation of LaCT~\cite{zhang2026testtime}. All fine-tuning runs on a single NVIDIA GH200 Grace Hopper with $96$\,GB of VRAM. The TTT module uses
a base window of $512$, a chunk size of $512$, and $32$ heads, with an $L_2$ inner objective, momentum with Muon orthogonalization, and a sigmoid output gate whose bias is initialized to $-3.0$. Following the two-stage LoLCATs recipe, we train attention transfer for $500$ steps at a learning rate of $2\times10^{-4}$, then fine-tune autoregressively for $400$ steps at
$1\times10^{-4}$ with rank-8 LoRA on the query, key, and value projections. Both stages use AdamW with gradient clipping at $1.0$ and an effective batch size of $8$, accumulated from microbatches of size $1$. Training maintains $28$ shared memories, of which $2$ are retained at deployment and additional system configurations are discussed in appendix~\ref{adp:expr}.

\begin{table}[t]
\caption{\textbf{Study on End-to-end latency (ms)} of \textsc{Anchor} variants compared
with the softmax-attention baselines (Llama and Mistral) across sequence lengths with the batch-size (B) of $19$. \textsc{Anchor} reduces end-to-end latency by up to $4.3\times$ relative to the baseline at 8K, and still runs at 16K, where both baselines exhaust memory (\textcolor{red}{OOM}) : Out-Of-Memory.}
\label{tab:anchor-efficiency}
\vskip 0.1in
\centering
\small
\setlength{\tabcolsep}{4pt}
\begin{tabular*}{\columnwidth}{@{\extracolsep{\fill}} l l cccc @{}}
\toprule
\multirow{2}{*}{\textbf{Model}} & \multirow{2}{*}{\textbf{Config.}} & \multicolumn{4}{c}{\textbf{E2E Latency Over Seq. Length (ms)}} \\
\cmidrule(l){3-6}
 & & 2K & 4K & 8K & 16K \\
\midrule
\multirow{4}{*}{LLaMa}
 & Baseline                 & $29164$ & $51049$ & $95387$ & \textcolor{red}{OOM} \\
\cmidrule(l){2-6}
 & \textsc{Anchor}-F        & $38028$ & $40719$ & $48471$ & $56516$ \\
 & \textsc{Anchor}-I        & $38770$ & $38333$ & $45219$ & $53775$ \\
 & \textbf{\textsc{Anchor}} & $\mathbf{19050}$ & $\mathbf{20854}$ & $\mathbf{22816}$ & $\mathbf{27915}$ \\
\midrule
\multirow{4}{*}{Mistral}
 & Baseline                 & $28872$ & $53100$ & $100300$ & \textcolor{red}{OOM} \\
\cmidrule(l){2-6}
 & \textsc{Anchor}-F        & $42465$ & $46782$ & $50600$ & $57700$ \\
 & \textsc{Anchor}-I        & $42318$ & $42000$ & $46700$ & $54300$ \\
 & \textbf{\textsc{Anchor}} & $\mathbf{20498}$     & $\mathbf{20200}$ & $\mathbf{23300}$ & $\mathbf{28600}$ \\
\bottomrule
\end{tabular*}
\end{table}

\paragraph{Models \& Datasets.} We evaluate \textbf{\textsc{Anchor}} on two open-weight models, Llama-3.1-8B and Mistral-7B. Three architectures are compared: \textsc{Anchor}-F, in which every sliding window has its own, non-shared TTT module; \textsc{Anchor}-I, which uses
cross-layer TTT modules at the anchor layers while retaining the interior TTT modules; and \textsc{Anchor}, which uses cross-layer TTT modules at the anchor layers and removes the interior modules entirely. Each is fine-tuned on the Long-Alpaca dataset~\cite{long-alpaca} for $500$ steps in Stage 1 and $400$ steps in Stage 2, and
the resulting checkpoints are used to evaluate language modeling and recall. For the empirical analysis, we additionally fine-tune \textsc{Anchor} on PG-19 and use it to study long-context retrieval through associative recall and layer-wise representational similarity across TTT modules on PG-19 and WikiText.

\subsection{Results and Discussion} \label{sec:results}
We evaluate the \textsc{Anchor}-series of models  against the baseline architectures along three axes: language modeling (Section~\ref{sec: lang_model}), inference efficiency, and long-context sequence modeling (Section~\ref{sec: seq_model_long}). Additionally, We  study how architectural and training choices affect performance, and what they imply for redesigning the architecture to extend the performance--efficiency trade-off along with generalization.

\subsubsection{Efficient Language Modeling} \label{sec: lang_model}

\begin{figure*}[t]
\centering
\includegraphics[width=0.90\textwidth]{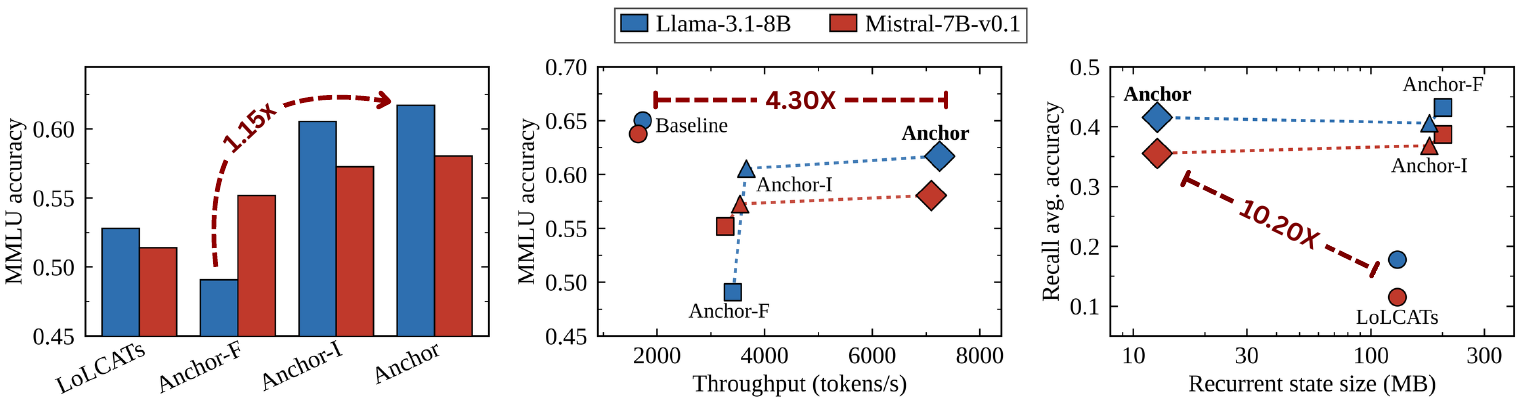}
\caption{\textbf{Study on Performance--efficiency trade-off.} \textsc{Anchor} evaluated on reasoning (MMLU) and recall (SWDE, FDA, SQuAD) benchmarks, together with end-to-end throughput and recurrent state size. \textbf{(a)} Reasoning accuracy of the \textsc{Anchor} series against LoLCATs; targeted placement of the cross-layer recurrent state improves reasoning. \textbf{(b)} Throughput (tokens/s) against reasoning accuracy; \textsc{Anchor} attains $4.3\times$ higher throughput than the baseline at comparable accuracy. \textbf{(c)} Recall accuracy against recurrent state size; \textsc{Anchor} improves recall over LoLCATs by 2.3--3.1$\times$ while using a 10.2$\times$ smaller state.}
\label{fig:mmlu}
\end{figure*}

\textbf{Efficient Common-sense reasoning.} We first establish the cost of linearization on general language modeling before evaluating long-context retrieval. We compare the \textsc{Anchor} series against the softmax-attention baselines (Llama and Mistral), their sliding-window variants, and LoLCATs linearization on six common-sense benchmarks covering sentence completion, which is illustrated in Table~\ref{tab:lm-eval}, physical and social reasoning, and science question answering. The sliding-window configurations provide the reference point: a $512$-token window alone remains within one point of full attention on both models, indicating that the inputs in these benchmarks largely fall within the window and that the long-term memory is only weakly exercised. These benchmarks are therefore informative about whether linearization degrades general language modeling, rather than about memory design. We observe no such degradation. All \textsc{Anchor} configurations remain within three points of their respective baselines, and the differences among the variants are sufficiently small that none is uniformly best across models, so accuracy alone does not distinguish them. They differ instead in cost: \textsc{Anchor} attains comparable accuracy with $12.6$\,MB of recurrent state, $16\times$ less than \textsc{Anchor}-F and $10\times$ less than LoLCATs.

\paragraph{Improved Recall Design.} We next evaluate long-context retrieval on recall benchmarks following~\cite{arora2025simplelinearattentionlanguage}, reported in Table~\ref{tab:lm-eval}. On Llama, \textsc{Anchor} more than doubles the average recall of LoLCATs while using $10\times$ less recurrent state. The margin is
largest on the tasks whose targets lie furthest outside the local window: on SWDE and FDA, LoLCATs scores below two points, against $49.9$ and $24.3$ for \textsc{Anchor}. A $512$-token sliding window carrying no persistent state
outperforms LoLCATs, indicating that capacity alone does not recover long-context recall; the determining factor is how the state is written and read. \textsc{Anchor} then comes within two points of \textsc{Anchor}-F while replacing the per-layer states of all $32$ layers with a single shared state per anchor group, a $16\times$ reduction, and the same ordering holds on Mistral. A gap to softmax attention remains, so anchor placement improves linearized recall without fully recovering it. We further observe that the recall gap between linearized hybrids and either the quadratic baseline or a hybrid pretrained from scratch persists regardless of long-context fine-tuning, \textit{which may be a consequence of the supervised fine-tuning used in stage-2}~\cite{zhou2026attention}. To separate the architecture from the linearization procedure, we train it from scratch on the MAD
synthetic tasks~\cite{poli2024mechanisticdesignscalinghybrid}, where \textsc{Anchor}-I matches or exceeds the Transformer (93.5 versus 90.9; Table~\ref{tab:mad}), notably on fuzzy recall, \textit{suggesting
that the recall gap arises from linearization rather than the architecture} (Appendix~\ref{apd:long_context}).

\begin{table}[t]
\centering
\caption{\textbf{Mechanistic architecture design (MAD) synthetic tasks.} Accuracy (\%) of
each architecture trained from scratch on the MAD suite. \textsc{Anchor}-I attains the
highest average and the largest gain on fuzzy recall.}
\label{tab:mad}
\small
\setlength{\tabcolsep}{4pt}
\renewcommand{\arraystretch}{1.15}
\resizebox{\columnwidth}{!}{%
\begin{tabular}{@{}l cccccc@{}}
\toprule
\textbf{Model} & \makecell{\textbf{IC} \&\\\textbf{Noisy}} & \textbf{Fuzzy} & \textbf{Mem.} & \makecell{\textbf{Sel.}\\\textbf{Copy}} & \textbf{Comp.} & \textbf{Avg.} \\
\midrule
Transformer       & $98.7$  & $55.7$ & $100.0$ & $99.9$ & $100.0$ & $90.9$ \\
$3:1$-Hybrid        & $96.8$  & $37.7$ & $99.9$  & $86.4$ & $100.0$ & $84.1$ \\
\midrule
\rowcolor{green!12}
\textbf{\textsc{Anchor}}-F & $100.0$ & $58.0$ & $100.0$ & $99.9$ & $100.0$ & $91.6$ \\
\rowcolor{green!12}
\textbf{\textsc{Anchor}}-I & $\mathbf{100.0}$ & $\mathbf{67.9}$ & $100.0$ & $99.9$ & $100.0$ & $\mathbf{93.5}$ \\
\bottomrule
\end{tabular}}
\end{table}

 \begin{figure*}[t]
\centering
\includegraphics[width=0.85\textwidth]{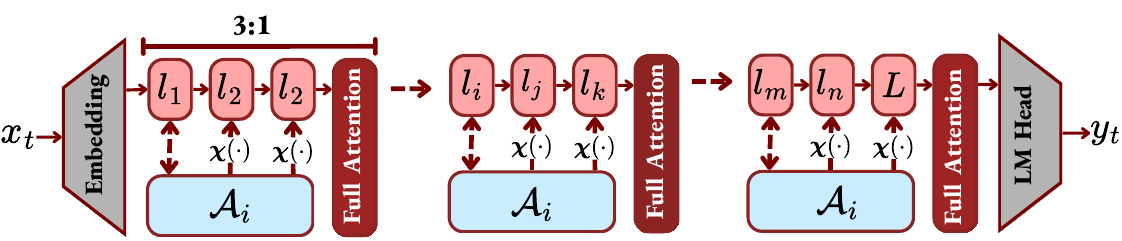}
\caption{\textbf{Hybrid architecture of \textsc{Anchor} and full self-attention.}
\textsc{Anchor} and full self-attention layers are interleaved at a $3{:}1$ ratio, so
that softmax attention accounts for 25\% of all layers. Within each group, the three
\textsc{Anchor} layers retain their own sliding-window attention for local context but
share a single recurrent state for long-range context: the leading layer writes the
state, and each layer reads it through its own learned read map $\chi_\ell(\cdot)$. The
subsequent full self-attention layer attends over the entire sequence, providing exact
retrieval at regular depth intervals, while the recurrent state per group remains
constant in sequence length.}
\label{fig:anchor_long}
\end{figure*}

\paragraph{Pareto-Optimal Architecture.}
Across all configurations, the \textsc{Anchor} series sustains a favourable efficiency--performance trade-off relative to both the quadratic baselines and prior linearized architectures. \textsc{Anchor} is the lightest architecture evaluated yet attains the highest MMLU score among the linearized variants, retaining $91$--$95\%$ of baseline accuracy and surpassing LoLCATs by $1.13$--$1.17\times$ consistently across backbones (Figure~\ref{fig:mmlu}(a)). On throughput, \textsc{Anchor}-F (\mksq{llamablue}, \mksq{mistralred}) and \textsc{Anchor}-I (\mktri{llamablue}, \mktri{mistralred}) already exceed the Llama~3-8B (\mkcirc{llamablue}) and Mistral-7B (\mkcirc{mistralred}) baselines at a context length of $8$K, and the deployment configuration \textsc{Anchor} (\mkdia{llamablue}, \mkdia{mistralred}) decodes $4.30\times$ faster than the baselines (Figure~\ref{fig:mmlu}(b)). On the recall benchmarks, \textsc{Anchor} maintains higher accuracy than LoLCATs with a $10.2\times$ smaller recurrent state (Figure~\ref{fig:mmlu}(c)). \textsc{Anchor} therefore \textit{lies on the Pareto frontier, preserving broad knowledge and recall at a fraction of the decoding cost and state size}, which establishes anchor placement as a practical basis for scaling linearized models.

\paragraph{E2E Accumulated Latency.}
We further measure wall-clock end-to-end latency (prefill plus decode) for the baselines (Llama-3.1-8B, Mistral-7B-v0.1) and the \textsc{Anchor} series in a batched serving setting on a single NVIDIA GH200 with $95.1$\,GiB of usable device memory. Each request comprises one prefill of $N$ prompt tokens followed by $512$ cached decode steps under greedy decoding with no EOS early exit. Prefill and decode are timed separately with device synchronization around each phase, and the reported latency includes the TTT inner-loop state update and the read-map overhead. We use a batch size of $19$, the largest the softmax baseline sustains at $8$K, to approximate concurrent multi-user requests, and vary the sequence length from $2$K to $16$K. All models run in BF16; each configuration is preceded by two excluded warm-up iterations, and we report the mean over $10$ runs. 
As shown in Table~\ref{tab:anchor-efficiency}, \textit{\textsc{Anchor} attains the lowest end-to-end latency among all architectures}, a $\approx\!4.3\times$ reduction relative to the baselines, and the series sustains low latency as the context length grows without exceeding the baselines' memory footprint.

\subsubsection{Long-Context Sequence Modeling} \label{sec: seq_model_long}

\paragraph{Linearization Recipe.} The recall gap between the linearized architecture and both
the quadratic baseline and the pretrained hybrid motivates a linearization recipe tailored for
long-context inference (Appendix~\ref{apd:long_context}). We linearize the backbone by attention
transfer (Stage~1) on LongAlpaca-12k~\cite{long-alpaca}, augmented with $4{,}000$ unpacked short
rows from alpaca-cleaned~\cite{alpaca}, at a maximum context length of $8192$. We then fine-tune
autoregressively (Stage~2) on a mixture of $80\%$ ReactiveAI passkey-retrieval data and $20\%$
synthetic needle-in-a-haystack data generated at $4$K and $8$K lengths.

\begin{table}[h]
\centering
\caption{\textbf{Evaluation on Long-Context Retrieval (RULER).} We evaluate \textsc{Anchor}-hybrid on RULER at sequence lengths from $4$K to $16$K tokens and compare it with the LLaMA-3-8B teacher and Lizard, a prior hybrid long-context linearization method.}
\label{tab:long-context}
\small
\setlength{\tabcolsep}{6pt}
\begin{tabular}{@{}l c ccc@{}}
\toprule
\textbf{Model} & \textbf{Attn. (\%)} & \textbf{4K} & \textbf{8K} & \textbf{16K} \\
\midrule
LLaMA-3-8B (teacher) & $100$ & $92.3$ & $90.5$ & $85.7$ \\
\midrule
Lizard (hybrid)      & $50$  & $92.5$ & $91.2$ & $\mathbf{85.2}$ \\
\rowcolor{green!15}
\textbf{\textsc{Anchor}-Hybrid}        & $\mathbf{25}$ & $\mathbf{98.3}$ & $\mathbf{97.3}$ & $82.9$ \\
\bottomrule
\end{tabular}
\end{table}

\paragraph{Long-context retrieval Benchmark.} We further evaluate a $3{:}1$ hybrid
configuration (Figure~\ref{fig:anchor_long}) in which every three \textsc{Anchor} layers are followed by one
vanilla softmax attention layer, so that softmax attention accounts for $25\%$
of all layers. This interleaving follows
Kimi-Linear~\cite{kimiteam2025kimilinearexpressiveefficient} and is intended to
narrow the gap between the linearized model and its pretrained teacher without reintroducing the full quadratic cost. We evaluate on RULER~\cite{hsieh2024rulerwhatsrealcontext} at sequence lengths from $4$K to
$16$K against the LLaMA-3-8B teacher ($100\%$ softmax attention) and
Lizard~\cite{Nguyen_2026} ($50\%$), as reported in Table~\ref{tab:long-context}. At $4$K and $8$K, \textsc{Anchor} attains the highest accuracy of the three configurations while using the least softmax attention, exceeding the teacher by $6.0$ and $6.8$ points, respectively. At $16$K the ordering reverses: accuracy falls to $82.9$, below both the teacher
($85.7$) and Lizard ($85.2$). All three models degrade as the context grows,
but the decline is steeper for \textsc{Anchor} ($15.4$ points, against $6.6$ for the teacher and $7.3$ for Lizard), indicating that the fine-tuning is introducing some level of bias towards attention along the near diagonal, which potentially limits the retrieval at the longer context. 

\begin{figure}[h]
  \centering
  \includegraphics[width=0.90\columnwidth]{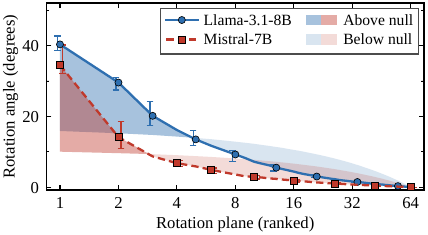}
  \caption{\textbf{The learned reader rotation is concentrated in a few planes.}
  Rotation angles of $Q$'s planes, ranked (log scale), at L31; median over
  32 heads, whiskers show the IQR. Shaded bands span from the learned profile to
  a random rotation with equal total squared angle (dark: learned above null;
  light: below). The learned map exceeds the null only in the leading planes
  (up to plane~5 for Llama-3.1-8B and plane~3 for Mistral-7B), and its top 5
  planes carry 82\% and 92\% of the squared rotation versus 26\% for the null.This holds in all 128 reader-layer heads of both backbones.}
  \label{fig:angle-profile}
\end{figure}

\subsubsection{Architectural Analysis and Discussion}
\paragraph{Analysis of the Learned Read-Map.} Our empirical analysis shows that the memory of previously generated tokens encoded in the TTT modules is similar across layers, but expressed in layer-specific bases (Figure~\ref{fig:sim_check}). This naturally occurring similarity, together with
the mismatch in basis, motivates \textsc{Anchor}, a cross-layer recurrent state that allows several short-term attention branches to share a single long-term memory. Within such a state, one designated layer updates the memory at every generated token, and the remaining layers read from it through a learned read-map $\chi_\ell(\cdot)$. We expected these maps to learn the layer-specific rotations required to express the shared memory in each reader's own basis. Each read-map is implemented as a linear projection whose dimensions match the
per-head readout dimension, and is trained in both stages under the outer-loop objective. The projections are initialized to the identity, so that every reader begins from the representation written by the writer layer and departs from it
only as far as the objective requires. The learned maps behave as anticipated, and their rotation proves to be concentrated in a small number of planes rather than spread across the readout dimension: at L31, the five largest planes account for $82\%$ of the squared rotation angle for Llama-3.1-8B and $92\%$ for
Mistral-7B, against $26\%$ for a random rotation of the same total magnitude, and the same concentration holds across all $128$ heads of the four reader layers in both backbones (Figure~\ref{fig:angle-profile}). \textit{The read-maps are
therefore effectively low-rank: a shared state can serve many layers through a correction that acts on only a few directions. }A formal treatment of the layer-specific read-map, together with the statistical evidence for this behavior, is given in Appendix~\ref{apd:readmap_long}.

\begin{figure}[h]
  \centering
  \includegraphics[width=\columnwidth]{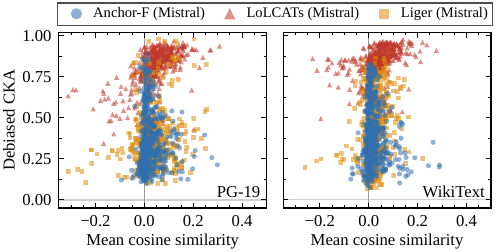}
  \caption{\textbf{Cross-layer similarity generalizes across linearization methods.}
Debiased CKA against mean cosine similarity for every pair of layers, on the
recurrent-state readouts of our TTT hybrid (Anchor-F), LoLCATs, and Liger on Mistral-7B over
identical PG-19 and WikiText inputs. In all three models, layers share information while
their cosine similarity remains small, indicating that this information is encoded in
different bases.}
  \label{fig:cka-generalize}
\end{figure}

\paragraph{Generalizing the Cross-Layer Recurrent State.} Having introduced a cross-layer recurrent state for the SWA--TTT hybrid, we ask: \textit{\textbf{can the same cross-layer design extend to other hybrid attention architectures?}} As a first step, we repeat the similarity analysis of Section~\ref{sec:analysis} on LoLCATs, which pairs a short sliding window with linear attention, and measure the representational similarity of its recurrent
states on both the corpus (PG-19 \& WikiText). The same phenomenon emerges on both. Recurrent states at different layers carry closely related information, with mean
debiased CKA of $0.79$ on PG-19 and $0.85$ on WikiText, and $93$--$100\%$ of layer pairs exceeding $0.5$, far above a shuffled-label floor below $0.004$. Yet their mean cosine similarity stays near zero ($0.04$ on both corpora), indicating that this shared information is encoded in different bases, as in our hybrid on the same inputs (mean CKA $0.36$ and $0.40$, mean cosine $0.02$; Figure~\ref{fig:cka-generalize}). That the pattern holds across three linearization
methods suggests that \textit{cross-layer redundancy is a general property of linearized recurrent states rather than an artifact of our architecture}, and that a shared state read through per-layer maps can extend to other hybrids. 

\section{Conclusion}
We propose \textbf{\textsc{Anchor}}, a cross-layer recurrent state for efficient long-context recall. A layer-wise ablation of the SWA--TTT hybrid shows that recall through the TTT modules concentrates in a few anchor layers while the interior layers can be removed, which we term the \textit{anchor effect}. Because states at different layers encode closely related information in different coordinate frames, \textsc{Anchor} keeps a single state written by one layer and read by many through learned, identity-initialized read maps. Across Llama-3.1-8B and Mistral-7B, it retains 91--95\% of baseline MMLU, improves recall over LoLCATs by 2.3--3.1$\times$, reduces the recurrent state by 16$\times$, and lowers end-to-end latency by up to 4.3$\times$, while running at 16K tokens where softmax baselines exhaust memory. The same structure emerges in LoLCATs, suggesting a general property of linearized recurrent states; position-aware state updates may further improve retrieval at the longest contexts.

\section{Acknowledgment}
 This work used DeltaAI at the National Center for Supercomputing Applications (NCSA) through allocation
 CIS$260333$ from the Advanced
 Cyberinfrastructure Coordination Ecosystem: Services \& Support (ACCESS) program, which is supported by
 U.S. National Science Foundation grants \#$2138259$, \#$2138286$, \#$2138307$, \#$2137603$, and \#$2138296$. 
 
\bibliographystyle{mlsys2025}
\bibliography{example_paper}

\section*{AI Usage Disclosure}
We used a state-of-the-art large language model (Claude, Anthropic)  as a programming assistant to integrate open-source packages, to support the implementation of the proposed architecture, and to write scripts for the planned experiments. We also used the LLM to edit the manuscript for clarity and presentation, including rephrasing text and preparing figure layouts and styling. All architectural design decisions, experimental design, and scientific conclusions are the authors' own. No reported data, results, or figure contents were generated by AI; every number and plot in the paper is derived from our own experiments. The authors take full responsibility for the content of this paper.

\clearpage

\appendix

\section{Preliminaries}
\label{apd:prelm}

\begin{figure*}[thb!]
  \centering
  \includegraphics[width=0.95\textwidth]{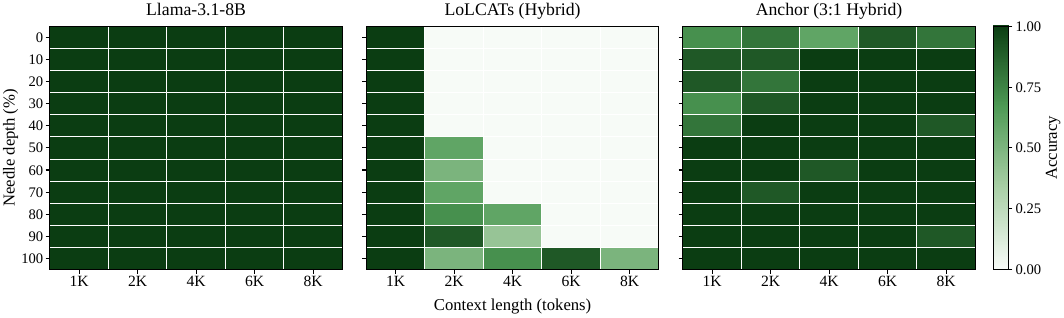}
  \caption{\textbf{Needle-in-a-haystack retrieval.} Accuracy across context lengths (1K--8K)
  and needle depths for Llama-3.1-8B, a LoLCATs hybrid ($18.75\%$~SoftMax Attention), and the \textsc{Anchor} 3:1 hybrid.
  \textsc{Anchor} reaches a mean accuracy of 0.95 (baseline 1.00), whereas the LoLCATs hybrid ($18.75\%$~SoftMax Attention)
  (0.33) retrieves reliably only at 1K or when the needle lies near the end of the context.}
  \label{fig:niah}
\end{figure*}

\begin{table}[t]
\centering
\caption{\textbf{Training Configuration of \textsc{Anchor}-F Architecture.} we trained the architecture separately for motivational study and main study with different dataset to understand to architecture behavior in different corpus and to maintain the fairness with other baselines. Context Length = $8192$ and Window Size = $512$.}
\label{tab:train-config}
\small
\setlength{\tabcolsep}{6pt}
\renewcommand{\arraystretch}{1.15}
\begin{tabular}{lcc}
\toprule
& \textbf{Stage 1} & \textbf{Stage 2} \\
\midrule
\multicolumn{3}{c}{\textbf{\textit{Motivational Study}}} \\
\midrule
\textbf{Dataset}          & \multicolumn{2}{c}{emozilla/pg19} \\
\textbf{Train / Val docs} & $200$ / $20$ & $200$ / $20$ \\
\textbf{Learning Rate (LR)}    & $2{\times}10^{-4}$ & $1{\times}10^{-4}$ \\
\textbf{Max steps}        & $400$ & $200$ \\
\textbf{Eval every }      & $25$  & $25$ \\
\textbf{Objective }       & MSE + CE & CE \\
\midrule
\multicolumn{3}{c}{\textbf{\textit{Main Experiments}}} \\
\midrule
\textbf{Dataset }         & \multicolumn{2}{c}{Yukang/LongAlpaca-12k} \\
\textbf{Train / Val docs} & $12{,}000$ / $100$ & $12{,}000$ / $100$ \\
\textbf{Learning Rate (LR)}   & $2{\times}10^{-4}$ & $1{\times}10^{-4}$ \\
\textbf{Max steps }       & $500$ & $400$ \\
\textbf{Eval every}       & $25$  & $100$ \\
\textbf{Objective}        & MSE + CE & CE \\
\bottomrule
\end{tabular}
\end{table}

\paragraph{Self-Attention.} For an input hidden state $\mathbf{x}_i$, the query,
key, and value are obtained as linear projections
$\hat{\mathbf{q}}_i = \mathbf{W}_Q^{\top}\mathbf{x}_i$,
$\hat{\mathbf{k}}_i = \mathbf{W}_K^{\top}\mathbf{x}_i$, and
$\hat{\mathbf{v}}_i = \mathbf{W}_V^{\top}\mathbf{x}_i$. The attention weights~\cite{vaswani2017attention}
$\boldsymbol{\alpha}_i$ form a distribution over the current query and all
preceding tokens,

\begin{equation}
    \alpha_{ij} = \frac{\exp\!\big(\hat{\mathbf{q}}_i^{\top}
    \hat{\mathbf{k}}_j / \sqrt{d_k}\big)}
    {\sum_{j'=0}^{i}\exp\!\big(\hat{\mathbf{q}}_i^{\top}
    \hat{\mathbf{k}}_{j'} / \sqrt{d_k}\big)},
\end{equation}

and the layer output is formed by aggregating the values under these weights
and projecting the result, $\mathbf{o}_i = \big(\sum_{j \leq i}\alpha_{ij}
\hat{\mathbf{v}}_j\big)\mathbf{W}_O$. Because the exponential similarity admits
no finite-dimensional factorization, the sum over $j$ cannot be accumulated into
a fixed-size state: every $(\hat{\mathbf{k}}_j, \hat{\mathbf{v}}_j)$ pair must
be retained, which is the origin of the KV cache and of its linear growth with
context.

\paragraph{Sliding Window Attention.} Windowed attention~\cite{beltagy2020longformerlongdocumenttransformer} bounds the number of
tokens retained in the KV cache by restricting each query to the $W$ most recent
positions, so that

\begin{equation}
\alpha_{ij} = \frac{\exp\!\big(\hat{\mathbf{q}}_i^{\top}
\hat{\mathbf{k}}_j/\sqrt{d_k}\big)}
{\sum_{j'=i-W+1}^{i}\exp\!\big(\hat{\mathbf{q}}_i^{\top}
\hat{\mathbf{k}}_{j'}/\sqrt{d_k}\big)},
\end{equation}

and $\alpha_{ij} = 0$ otherwise. The cache therefore holds at most $W$
key--value pairs per layer, and the per-token retrieval cost becomes
independent of the sequence length and with the significant performance degradation in longer-context.

\paragraph{Linear Attention.} Self-attention can be linearized by replacing the
softmax with a separable kernel feature map $\phi$. The resulting linear
attention~\cite{katharopoulos2020transformersrnnsfastautoregressive} is

\begin{equation}
\label{eq:linattn}
\mathbf{o}_i =
\frac{\sum_{j=0}^{i}\phi(\hat{\mathbf{q}}_i)\,
\phi(\hat{\mathbf{k}}_j)^{\top}\hat{\mathbf{v}}_j}
{\sum_{j=0}^{i}\phi(\hat{\mathbf{q}}_i)\,\phi(\hat{\mathbf{k}}_j)^{\top}}.
\end{equation}

By the associativity of matrix multiplication, the sums over $j$ can be factored
out of the query, giving $\mathbf{o}_i = \phi(\hat{\mathbf{q}}_i)\mathbf{S}_i /
\phi(\hat{\mathbf{q}}_i)\mathbf{Z}_i$, where
$\mathbf{S}_i = \sum_{j=0}^{i}\phi(\hat{\mathbf{k}}_j)^{\top}\hat{\mathbf{v}}_j
\in \mathbb{R}^{d_k \times d_v}$ is the recurrent state and
$\mathbf{Z}_i = \sum_{j=0}^{i}\phi(\hat{\mathbf{k}}_j)^{\top} \in
\mathbb{R}^{d_k}$ is the normalization factor.  Both accumulate incrementally and are independent of sequence length, so the per-token cost no longer grows with context. Expressiveness then depends on the
choice of $\phi$, and recall on the capacity of the state~\cite{arora2025simplelinearattentionlanguage}.

\section{Experimental Setup \& Definitions}
\label{adp:expr}
We introduces \textbf{\textsc{Anchor}}-series of linearized LLM architectures with the various level of efficiency-performance trade-offs. The Architectural definition and their training configurations can be illustrated as, 

\paragraph{\textsc{Anchor}-Full (or) \textsc{Anchor}-F.} \textsc{Anchor}-F is a linearized architecture that combines Sliding Window Attention (SWA) with a Test-Time Training (TTT) module to form a hybrid attention, in which each SWA layer maintains its own TTT module for long-context recall. This configuration is used throughout the motivational
experiments, both for the layer-wise study of recall and for the analysis of cross-layer similarity between recurrent states.
Table~\ref{tab:train-config} reports its training configuration. In both the
motivational study and the main experiments we use AdamW with a weight decay of
$0.01$ and BF16 precision.

\paragraph{\textsc{Anchor}-I and \textsc{Anchor}.} Motivated by the cross-layer similarity and the sparsity of recall across TTT
layers, we introduce \textbf{\textsc{Anchor}}-I, a training-specific configuration in which layers with high representational similarity are replaced by a shared recurrent state while the interior TTT layers are retained. The
interior layers are kept only during training, where they prevent the gradient
instability observed when \textsc{Anchor} is trained without them. \textsc{Anchor}-I uses the same training configuration as the main experiments in Table~\ref{tab:train-config}. \textsc{Anchor} is then the lightweight runtime
configuration, in which the interior TTT layers are removed for efficient
inference.

\section{Recurrent State with Layer-Specific Transformation} \label{apd:readmap_long}

\textbf{Approximating Linear Attention with Neural Networks.} Following prior
work~\cite{liu2026testtime}, the long-term memory can be approximated by a neural network
trained online on the association between keys and values, an objective termed the attentional bias. This yields a more expressive form of linear attention whose recurrent
state $\mathbf{S}_i$ is updated by a gradient step,

\begin{equation}
  \mathbf{S}_i = \mathbf{S}_{i-1} - \eta\, \nabla_{\mathbf{S}}\,
  \mathcal{L}\!\left(\mathbf{S}_{i-1}(\mathbf{k}_i),\, \mathbf{v}_i\right),
\end{equation}
where $\eta$ is the inner learning rate and
$\mathcal{L}(\hat{\mathbf{v}}, \mathbf{v}) = \tfrac{1}{2}\lVert \hat{\mathbf{v}} - \mathbf{v} \rVert_2^2$;
the state is read as a function of the query, $\mathbf{o}_i = \mathbf{S}_i(\mathbf{q}_i)$. Motivated by the layer-specific
transformations we observe across linearized layers (Section~\ref{sec:analysis}), we generalize the readout of layer $\ell$ to

\begin{equation}
  \mathbf{o}_i^{(\ell)} = \boldsymbol{\chi}_\ell\!\left(\mathbf{S}_i(\mathbf{q}_i^{(\ell)})\right)
  \in \mathbb{R}^{d},
\end{equation}
where $\boldsymbol{\chi}_\ell$ is a layer-specific read map, initialized to the identity
$\mathbf{I} \in \mathbb{R}^{d \times d}$ so that each layer initially reads its state
exactly as written. By polar decomposition, any such map factors as
$\boldsymbol{\chi}_\ell = \mathbf{U}_\ell \mathbf{P}_\ell$, where $\mathbf{U}_\ell$ is orthogonal, a rotation of the basis, and $\mathbf{P}_\ell$ is symmetric positive semi-definite, a stretch along its principal axes. To compare layers, note that each layer writes its state in its own basis, so the
identity read map of a per-layer model is the identity \emph{relative to that layer}: for two layers $\ell_m$ and $\ell_n$, $\mathbf{I}_{\ell_m}$ and $\mathbf{I}_{\ell_n}$ act
in different coordinate frames. Expressed in a common frame, the two are related by a
transformation
\begin{equation}
  \mathbf{T}_{\ell_m \to \ell_n} = \mathbf{U}_{mn}\,\mathbf{P}_{mn} \neq \mathbf{I},
\end{equation}

Hence, to recover a state written in the basis of layer $\ell_m$, the read map of layer $\ell_n$ must realize an anisotropic stretch $\mathbf{P}_{mn}$ followed by a
rotation $\mathbf{U}_{mn}$. The cross-layer analysis (Figure~\ref{fig:sim_check}) supports $\mathbf{T}_{\ell_m\to\ell_n} \neq \mathbf{I}$: states of different layers attain high
debiased CKA, which is invariant to rotation, while their mean cosine similarity is near zero, i.e., the layers encode related information in different coordinates.

\begin{figure*}[p]
  \centering
  \begin{subfigure}{\textwidth}
    \centering
    \includegraphics[width=0.85\textwidth]{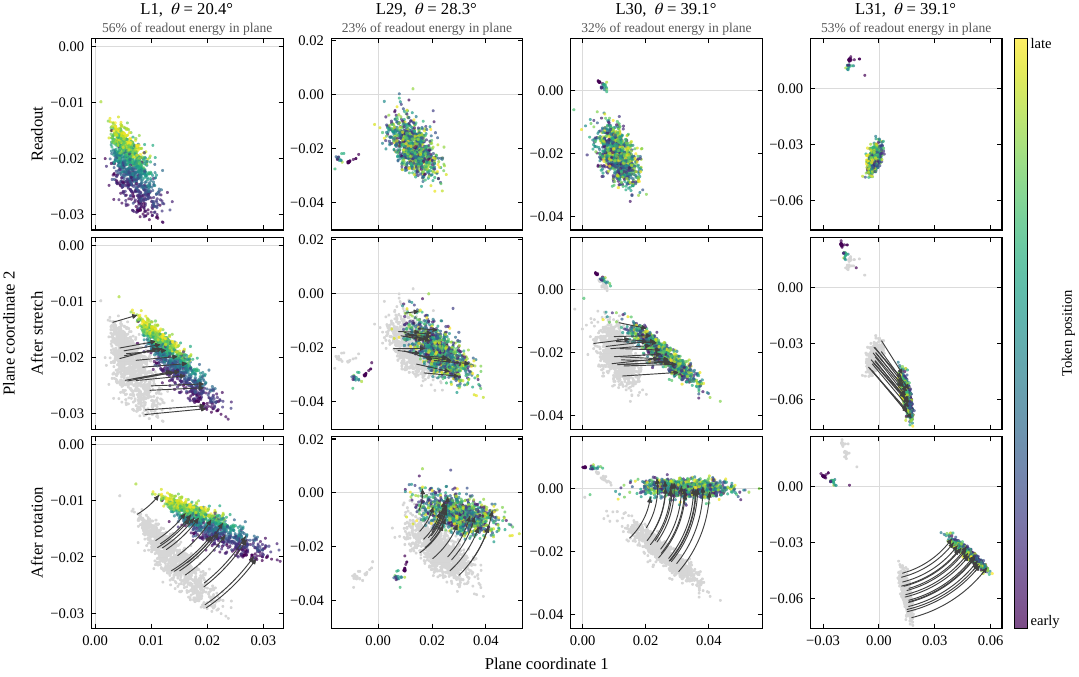}
    \caption{Llama-3.1-8B}
    \label{fig:readmap-llama}
  \end{subfigure}

  \vspace{0.6em}

  \begin{subfigure}{\textwidth}
    \centering
    \includegraphics[width=0.85\textwidth]{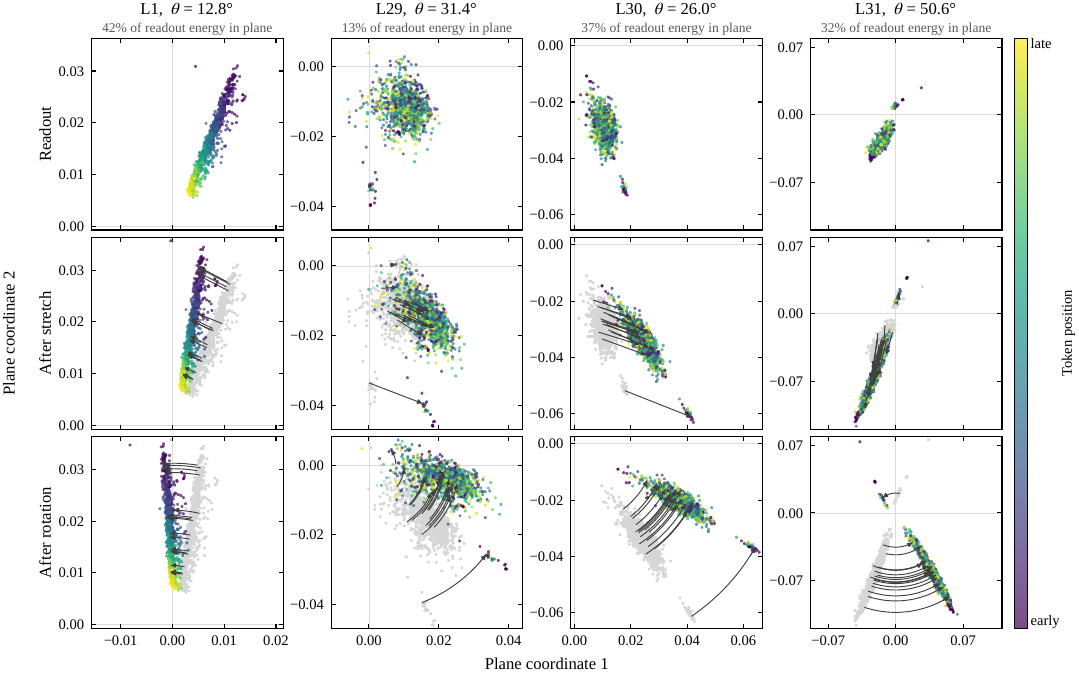}
    \caption{Mistral-7B-v0.1}
    \label{fig:readmap-mistral}
  \end{subfigure}

  \caption{\textbf{Geometry of the learned read maps.}
  Each column is one reader layer, and the rows show the readout as it passes through
  the two stages of the read map $Q$: a stretch followed by a rotation. Points are readout
  vectors projected onto the layer's dominant rotation plane and coloured by position in
  the sequence; arcs trace the rotation step, which preserves the radius within the plane.
  Panel titles give the rotation angle $\theta$ in that plane and the share of readout
  energy it carries. Both backbones follow the same pattern, with the rotation concentrated
  in a few planes (Figure~8). Head 0 of 32; axes are scaled per layer.}
  \label{fig:readmap}
\end{figure*}

\textbf{Cross-Layer Recurrent State with a Learnable Transformation.} A practical cross-layer recurrent state $\mathcal{A}_i$, shared by several short-range attention layers, needs each reading layer to realize the transformation in Eq.~(7), e.g., $\mathbf{U}_{mn}\mathbf{P}_{mn}$ for a state written by layer $\ell_m$ and read by layer $\ell_n$. We parameterize this transformation as a learnable bias-free linear map. For a cross-layer state led by layer $\ell_m$, the transformation for layer $\ell_n$ is

\begin{equation}
  \chi_{\ell_n}(\mathbf{x}) = \mathbf{M}_{\theta_n}\mathbf{x}
  \;\;\text{with}\;\; \mathbf{M}_{\theta_n} \to \mathbf{T}_{\ell_m \to \ell_n},
\end{equation}
where $\mathbf{x} = \mathcal{A}_i(\mathbf{q}_i^{(\ell_n)})$ is the raw readout of the shared state and $\mathbf{M}_{\theta_n} \in \mathbb{R}^{d\times d}$ is initialized to
$\mathbf{I}$; the writer layer $\ell_m$ keeps $\chi_{\ell_m} = \mathbf{I}$. Since every square matrix admits a polar decomposition, a single bias-free linear map can represent
both the anisotropic stretch $\mathbf{P}_{mn}$ and the rotation $\mathbf{U}_{mn}$. Figures~\ref{fig:readmap-llama} and~\ref{fig:readmap-mistral} visualize the learned read
maps on both backbones. Their departure from the identity is low-rank: the rotation is concentrated in a few planes, with the top five carrying 82\% (Llama) and 92\% (Mistral) of the rotation. 

\section{Long-Context Configuration} \label{apd:long_context}

We construct a hybrid architecture that
interleaves vanilla softmax ~\cite{chen2026hybridlinearattentionright, du2026nativehybridattentionefficient, nvidia2026nemotron3ultraopen} with the \textsc{Anchor} cross-layer
recurrent state at a $3{:}1$ ratio: every three \textsc{Anchor} layers, which
share a single recurrent state, are followed by one softmax layer, so that
softmax attention accounts for $25\%$ of all layers
(Figure~\ref{fig:anchor_long}). The design narrows the gap between the
linearized model and its pretrained teacher through minimal interleaving,
without reintroducing the full quadratic cost. This hybridization makes the
model competitive on long-context retrieval, as shown in
Figure~\ref{fig:niah} and Table~\ref{tab:long-context}.

\paragraph{On Pre-training Scale. }To assess the \textsc{Anchor} series under from-scratch pretraining rather than
distillation, we additionally evaluate it on the mechanistic architecture design
(MAD) suite~\cite{poli2024mechanisticdesignscalinghybrid}, reported in Table~\ref{tab:mad}.
\textsc{Anchor}-I attains the highest average accuracy ($93.5$), ahead of both
the Transformer ($90.9$) and \textsc{Anchor}-F ($91.6$). Memorization,
compression, and selective copying saturate across variants, so the comparison
is decided by the two harder tasks. Both \textsc{Anchor} configurations reach
$100.0$ on in-context recall with noise, against $98.7$ for the Transformer, and
fuzzy recall separates the architectures most clearly, with \textsc{Anchor}-I at
$67.9$ against $55.7$. The plain $3{:}1$ hybrid trails at $84.1$ overall and
falls to $37.7$ on fuzzy recall, indicating that the gain comes from the
cross-layer recurrent state rather than from interleaving alone. Taken together,
these results suggest that the cross-layer state is not merely a cost-efficient
substitute obtained through distillation, but a mechanism that improves
associative retrieval when trained from scratch.

\label{app:window}

\end{document}